\documentclass{article} 
\usepackage{iclr2026_conference,times}

\usepackage{amsmath,amsfonts,bm}

\def\eqref#1{equation~\ref{#1}}

\def\1{\bm{1}}

\DeclareMathAlphabet{\mathsfit}{\encodingdefault}{\sfdefault}{m}{sl}
\SetMathAlphabet{\mathsfit}{bold}{\encodingdefault}{\sfdefault}{bx}{n}

\newcommand{\KL}{D_{\mathrm{KL}}}

\usepackage[table]{xcolor}
\usepackage[acronym, nonumberlist, nogroupskip]{glossaries}
\usepackage{graphicx}
\usepackage{hyperref}
\usepackage{url}
\usepackage{enumerate}
\usepackage{enumitem}
\usepackage{booktabs}
\usepackage{algpseudocode}
\usepackage{algorithm}
\usepackage{wrapfig}
\usepackage{multirow} 
\newcommand{\algname}{{{LookUM}}}

\definecolor{cornellred}{rgb}{0.7, 0.11, 0.11}
\definecolor{cadmiumgreen}{rgb}{0.0, 0.42, 0.24}
\definecolor{aliceblue}{rgb}{0.91, 0.94, 0.97}
\definecolor{darkblue}{rgb}{0.83, 0.89, 0.97}
\definecolor{Red7}{rgb}{0.941, 0.243, 0.243}
\definecolor{Green7}{RGB}{55, 178, 77}
\definecolor{Blue9}{rgb}{0.098,0.3,0.9}

\hypersetup{
  linkcolor = cornellred,
  citecolor  = cadmiumgreen,
  colorlinks = true,
}
\title{Lookahead Unmasking Elicits Accurate \\ Decoding in Diffusion Language Models}

\author{Sanghyun Lee$^1$\thanks{This work was done during an internship at KRAFTON.} , Seungryong Kim$^1$, Jongho Park$^2$, Dongmin Park$^{3}$\thanks{Corresponding author.} \\
$^1$KAIST \quad $^2$University of California, Berkeley\quad$^3$ KRAFTON\\
\texttt{\{lsh83210,seungryong.kim\}@kaist.ac.kr\quad  jjhpark@berkeley.edu\quad} \\ \texttt{dongmin.park@krafton.com}}

\newacronym{dlm}{DLM}{Diffusion Language Model}
\newacronym{mdm}{MDM}{Masked Diffusion Model}

\iclrfinalcopy 
\begin{document}

\maketitle

\begin{abstract}
Masked Diffusion Models (MDMs) as language models generate by iteratively unmasking tokens, yet their performance crucially depends on the inference-time order of unmasking.
Conventional methods such as confidence-based sampling are short-sighted, focusing on local optimization while neglecting additional test-time computation and allowing early decoding errors to cascade. We propose \textbf{Lookahead Unmasking (\algname{})}, which addresses these concerns by reformulating sampling as \textit{path selection} over all possible unmasking orders without the need for an external reward model. 
Our framework couples (i) a path generator that proposes paths by sampling from pools of unmasking sets with (ii) a verifier that computes the uncertainty of the proposed paths and performs importance sampling to subsequently select the final paths.
Empirically, erroneous unmasking measurably inflates sequence-level uncertainty, and our method exploits this to avoid error-prone trajectories. 
We validate our framework across six benchmarks, such as mathematics, planning, and coding, and demonstrate consistent performance improvements.
\algname{} requires only two to three paths to achieve peak performance, demonstrating remarkably efficient path selection. The consistent improvements on both LLaDA and post-trained LLaDA 1.5 are particularly striking: base LLaDA with \algname{} rivals the performance of RL-tuned LLaDA 1.5, while \algname{} further enhances LLaDA 1.5 itself—showing that uncertainty-based verification provides orthogonal benefits to reinforcement learning and underscoring the versatility of our framework.

\end{abstract}

\section{Introduction}
Following the remarkable success of diffusion models in continuous domains such as image and video~\citep{song2020score, rombach2022high, ho2022video}, diffusion language models (DLMs) have recently emerged as a promising and efficient alternative to autoregressive language models (ARMs) for discrete sequence generation~\citep{austin2021structured,sahoo2024simple,lou2023discrete}. Notably, state-of-the-art DLMs, including LLaDA~\citep{nie2025large} and Dream~\citep{ye2025dream}, adopt the form of masked diffusion models (MDMs), where generation proceeds through iterative unmasking over sequences of masked tokens. Recent advances have further enhanced these models through reinforcement learning-based fine-tuning, as demonstrated in LLaDA 1.5~\citep{zhu2025llada} and d1~\citep{zhao2025d1}, which employ reward-based optimization to improve reasoning capabilities.

The primary distinction between ARMs and DLMs lies in the order of token unmasking. Unlike ARMs, DLMs possess the flexibility to arbitrarily select the positions of tokens to be unmasked, in addition to predicting token values. For instance, in tasks such as planning~\citep{ye2024beyond} and infilling~\citep{gong2024scaling}, where bidirectional rather than left-to-right unmasking is expected to be advantageous, DLMs demonstrate clear superiority over ARMs. This highlights that the order of unmasking is not merely an architectural detail but a critical factor in achieving strong performance.
\begin{figure*}[ht!]
    \centering
    \includegraphics[width=0.99\linewidth]{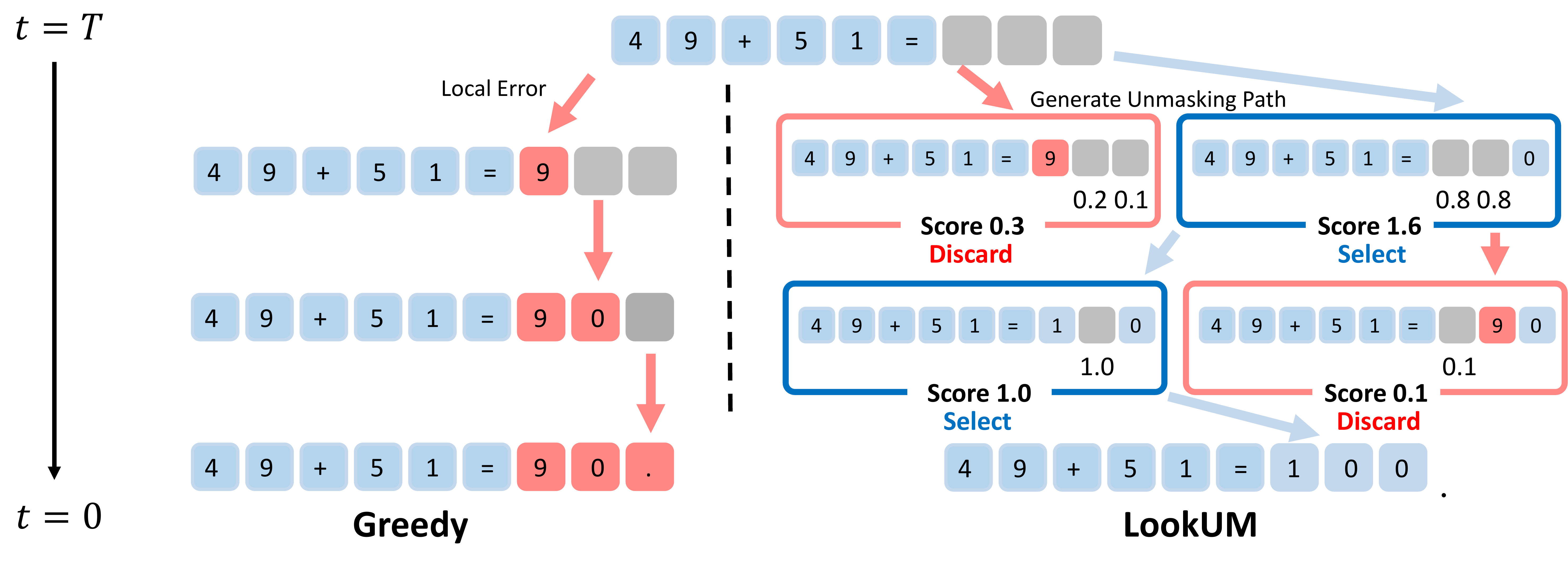}
    \vspace{-5pt}
    \caption{\textbf{Standard unmasking vs. \algname{} in discrete diffusion models.} During the denoising process of unmasking from timestep $T$ to $0$, greedy approaches often select the position with the highest token-level certainty which can lead to an incorrect unmasking order and result in local errors (\textcolor{red}{\textbf{red}}). In contrast \algname{} generates candidate unmasking paths and leverages a verifier to select those that avoid local errors and recover the correct sequence (\textcolor{blue}{\textbf{blue}}). }
    \label{fig:greedy_lookum}
\end{figure*}
Existing approaches to unmasking predominantly adopt heuristic strategies that rely on token-level model predictions to compute certainty measures (e.g., negative entropy, confidence, margin), which are then used to greedily select the next position to unmask~\citep{chang2022maskgit,huang2025pc,kim2025train,koh2024plm,ben2025accelerated}. Though lightweight and simple, these methods inherently reflect confidence in local token distributions rather than capturing sequence-level dependencies that govern the overall unmasking trajectory.
As a result, they often disregard global consistency and are prone to irreversible local errors.
As illustrated in Figure~\ref{fig:greedy_lookum}, a single incorrect unmasking decision (e.g., prematurely revealing a digit in arithmetic) can cascade through subsequent steps, making recovery impossible. We empirically observe in our experiments in Section~\ref{localerror}.

These limitations highlight the need for unmasking strategies that can identify potential local errors and steer the unmasking process toward paths that avoid them. To this end, we propose a framework that reformulates sampling as a path selection problem in the space of possible unmasking orders, where a verifier directs the process by detecting local errors and steering it toward more reliable paths. Building on this framework, we develop \textbf{Lookahead Unmasking (\algname{})}, which operates in a \textit{fully unsupervised} manner by leveraging a certainty measure over sequences as a verifier. 

Our method achieves reliable error avoidance with only a 2-3× computational cost. It evaluates candidate paths through a verifier and selects those that best preserve sequence-level coherence. In reasoning benchmarks, \algname{} consistently outperforms greedy unmasking, improving performance by up to 4 points on HumanEval and GSM8K, and 8 points on MBPP. Consistent gains in both LLaDA and its RL-tuned variant LLaDA 1.5 demonstrate that our method provides complementary benefits to the model training pipeline.

Overall, our contributions can be summarized as:
\begin{itemize}[leftmargin=10pt]
\item We reformulate unmasking as path selection and introduce \algname{}, which uses sequence-level uncertainty to guide generation.
\item \algname{} reduces local error rates by $10\%$ and improves accuracy by 4-8 points on reasoning benchmarks with only 2-3× overhead.
\item The method requires no external models and complements both base training and RL optimization, as shown on LLaDA 1.5.
\end{itemize}

\section{Background}

\paragraph{Masked Diffusion Language Models.}
A widely adopted family of discrete diffusion models for language is the \emph{masked diffusion} framework~\citep{austin2021structured,lou2023discrete,sahoo2024simple}, which reconstructs a text sequence from fully masked tokens. The process is formulated over timesteps $t\in [0,T]$, comprising (1) a forward noising process that transforms a token sequence $x_0$ into masked tokens, and (2) a reverse denoising process that reconstructs the sequence from the masked token $x_T$. Formally, let $\theta$ denote a DLM operating on sequences of length $L$. At each timestep $t$, the model handles $x_t = (x_t^1, x_t^2, \ldots, x_t^L)$, where each token $x_t^i$ belongs to a vocabulary $\mathcal{V} = \{1, 2, \ldots, m\}$, with the special symbol $m$ indicating the mask token. During denoising, the model sequentially determines the order of token unmasking.

\paragraph{Decoding process.} The decoding progresses through iterative denoising steps, gradually converting the fully masked sequence $x_T$ into a complete text sequence $x_0$. For each denoising step $t$, let $\mathcal{M}_t := \{ i \in \{1,\ldots,L\} \;:\; x_t^i = m \}$ be a \emph{masked index set} and $p_\theta^i(\cdot \mid x_t)$ be a predicted categorical distribution over $\mathcal{V}$ for each position $i$. Then, the objective of unmasking is to select $b_t$ indices to unmask from the currently masked positions $i \in \mathcal{M}_t$, based on $p^i_\theta(\cdot \mid x_t)$. That is, \emph{unmasking index set} $\mathcal{U}_t$ is selected and used to update the next masked index set, such as $\mathcal{M}_{t-1} = \mathcal{M}_t \setminus \mathcal{U}_t$.

In principle, token selection could be random, but the lack of explicit training constraints on the transition dynamics of the denoising process often leads to degraded performance~\citep{ben2025accelerated}. Therefore, practical implementations employ \emph{greedy} unmasking strategies, which consistently outperform random unmasking. Formally, with a scoring function $\sigma$ over token probability $p_\theta^i$, the unmasking index set is obtained as 
\begin{equation}
\label{eq:unmasking_index_set}
\mathcal{U}_t = \{i\in\mathcal{M}_t | ~ \text{rank}(\sigma^i_t) \leq b_t\},
\end{equation}
where $\sigma^i_t = \sigma(p^i_\theta(\cdot|x_t))$ is the score for token probability at position $i$ and $\text{rank}(\cdot)$ is its ranking over $\mathcal{M}_t$. Existing score functions include: (1) \textbf{Confidence}~\citep{chang2022maskgit}, $\sigma_t^i = p_t^{(i,1)}$; 
(2) \textbf{Margin}~\citep{kim2025train}, $\sigma_t^i = p_t^{(i,1)} - p_t^{(i,2)}$; 
(3) \textbf{Negative Entropy}~\citep{koh2024plm}, $\sigma_t^i = -H(p_\theta^i(\cdot \mid x_t))$ with $H(p_\theta^i) = -\sum_r p_t^{(i,r)} \log p_t^{(i,r)}$, where $p_t^{(i,r)} := p_\theta^i(v_t^{(i,r)} \mid x_t)$ is the probability of the token $v$ with $r$-th highest score over $\mathcal{V}$.

\section{Lookahead Unmasking for Diffusion Language Models}
This section introduces ~\algname{} for DLMs. We begin by illustrating how local errors arise in greedy unmasking and how sequence-level certainty enables effective path correction. We then reformulate unmasking as a path selection problem guided by certainty-based verifiers, and conclude with the algorithmic design and analysis showing significant performance improvements at low computational cost.

\subsection{Escaping Local Error Paths with Lookahead Unmasking}
\label{localerror}
As the model tends to unmask tokens only near already unmasked ones~\citep{gong2025diffucoder}, greedy unmasking often leads to errors that cannot be recovered as local error recovery is impossible~\citep{huang2025pc}.
Unmasking incorrect tokens induces local errors, including mathematical errors (digit misplacement, arithmetic mistakes), coding errors (variable reference errors, missing brackets or syntax), and general errors (illogical reasoning, grammatical inconsistencies). Once such errors occur, the greedy strategy has no mechanism to recover, and the generation becomes locked into an erroneous path.

\begin{wraptable}{r}{0.65 \linewidth}
\centering
\vspace{-20pt}
\caption{\textbf{Entropy and Confidence under Local Errors.} Entropy and confidence metrics across arithmetic operations (subtraction, addition, multiplication) demonstrate increased uncertainty when local errors occur versus correct predictions.}
\label{tab:uncertainty}
\vspace{10pt}
\begin{tabular}{lcccc}
\toprule
& \multicolumn{2}{c}{Correct} & \multicolumn{2}{c}{Error} \\
\cmidrule(lr){2-3} \cmidrule(lr){4-5}
& Entropy$\downarrow$ & Confidence$\uparrow$ & Entropy$\downarrow$ & Confidence$\uparrow$ \\
\midrule
Sub. &\textbf{0.24} & \textbf{0.97} & 1.69 & 0.82 \\
Add. & \textbf{0.53} & \textbf{0.93} & 1.82 & 0.82 \\
Mul. & \textbf{0.30} & \textbf{0.96} & 1.61 & 0.83 \\
\bottomrule
\end{tabular}

\label{tab:art}
\vspace{-1em}
\end{wraptable}

To address this limitation, we ground our approach in two key observations: (1) local errors propagate by increasing model uncertainty in subsequent predictions, and (2) paths with higher uncertainty exhibit greater susceptibility to local errors. These insights motivate \algname{}, which evaluates path-level certainty to selectively avoid error-prone trajectories while preserving sequence-level coherence.
\begin{figure*}[t!]
    \centering
    \includegraphics[width=\linewidth]{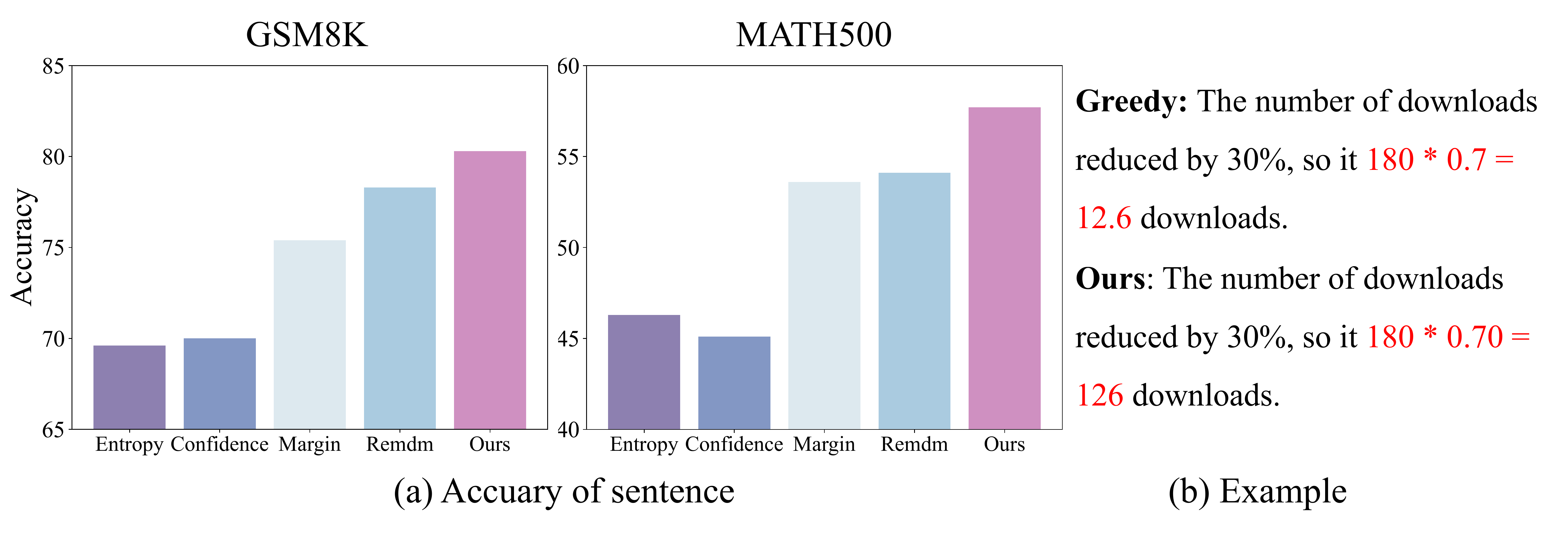}
    \vspace{-25pt}
    \caption{\textbf{Local Error Compare and Example.} (a) Sentence-level accuracy on GSM8K and MATH500, showing our method achieves approximately 10$\%$ lower error rates than baselines. (b) Example of greedy unmasking producing a computational error (180 × 0.7 = 12.6) while our method generates the correct result (126).}
\label{fig:local_error}
\end{figure*}

\paragraph{Local error increase model uncertainty.}
We evaluate the impact of local errors on model predictions using the Arithmetic dataset~\citep{brown2020language}, which contains 2,000 simple arithmetic problems. For each sequence position, we compare the subsequent predictions of the model under two conditions: (1) when the model produces the correct token, and (2) when a local error is deliberately introduced at the same position. We then measure the average entropy and confidence of the outputs generated in both settings. As shown in Table~\ref{tab:art}, the presence of local errors consistently increases the uncertainty of prediction, reflected by higher entropy values and lower confidence scores. This suggests that local errors propagate instability in the decoding process, ultimately making the model less reliable in its subsequent predictions.

\paragraph{Avoiding Uncertain Paths.}
In the previous analysis, we showed that local errors increase prediction uncertainty and may destabilize subsequent reasoning. Motivated by this observation, we propose Lookahead Unmasking (\algname{}), which exploits uncertainty as a signal for guiding the denoising process. The core idea is simple: instead of unmasking tokens in a fixed order, we perform a lookahead search over possible reasoning paths and dynamically adjust unmasking positions according to their certainty.

To demonstrate the effectiveness of this approach, we compare LookUM against baseline methods by measuring local error rates on MATH500~\citep{lightman2023let} and GSM8K~\citep{cobbe2021training}, using GPT-4o~\citep{openai2024gpt4} to verify the correctness of each reasoning step. As shown in Figure~\ref{fig:local_error}, LookUM achieves a consistent reduction of nearly 10$\%$ in local error rates compared to baseline methods, confirming that uncertainty-guided path selection effectively avoids error-prone trajectories. The complete algorithmic framework is presented in Section~\ref{lookaheadalg}, with implementation details in Appendix~\ref{expdetail}.

\subsection{Lookahead Unmasking}
\label{lookaheadalg}
We reformulate the unmasking procedure in DLMs as a path-selection problem over possible unmasking orders. Our framework consists of two components: a \emph{path generator} that proposes candidate unmasking sets, and an \emph{uncertainty-based verifier} that evaluates their reliability. This design transforms the myopic token-level decisions of existing methods into a lookahead mechanism that considers the downstream consequences of unmasking choices.

At each sampling step $t$, \emph{path generator} $G$ leverages the predictive distribution of the model $p_\theta(\cdot \mid x_t)$ together with the available budget $B_t$ to construct a candidate path, that is, an unmasking set $\mathcal{U}_t \subseteq \mathcal{M}_t$. This set specifies which positions are to be unmasked in the current step. The candidate path is then evaluated by the \emph{uncertainty-based verifier} $V$, which, for each position $i \in \mathcal{U}_t$, considers the transformed state where $x_t^i$ is sampled from its token distribution $p_\theta^i(\cdot|x_t)$. By analyzing the one-step-ahead dynamics under this hypothetical unmasking, the verifier assigns an uncertainty score that quantifies the expected consistency of the candidate path with future model predictions. The verifier can be regarded as a reward model that guides the selection of the path. Within this unifying perspective, unmasking can be interpreted as a reward alignment process, allowing its integration with existing methodologies to achieve reward alignment in sampling procedures.

\paragraph{Path Generator.}
Given the state $x_t$, the path generator returns a set of indexes
\[
G(p_\theta^i(\cdot|x_t),B_t,\mathcal{P}_t) = \mathcal{U}_t, \quad \mathcal{U}_t\subseteq \mathcal{M}_t, \; |\mathcal{U}_t^{(k)}| = B_t,
\]
constructed deterministically or stochastically from a proposal distribution. Deterministic generation can thus be viewed as the conventional top-$B_t$
 selection, resulting in a greedy choice of tokens, whereas in the stochastic case, unreliable outcomes may arise if proposals are drawn from low-certainty positions; therefore, the generator must instead sample from a high-certainty pool $\mathcal{P}_t$ with size $|\mathcal{P}_t|=N$, motivating us to propose several alternative constructions of pool as follows:
\begin{itemize}[leftmargin=1.5em]
    \item \textbf{N-best pooling:}  construct a candidate pool by selecting the top-$N$ tokens according to a certainty measure. 
The certainty measure can be defined using standard strategies such as entropy, confidence, or margin, allowing flexible
adaptation to different evaluation criteria.
    
    \item \textbf{Certainty filtering pooling:} include only tokens whose predictive 
    probability exceeds a pre-defined threshold $\tau$, ensuring proposals are sampled 
    from high-certainty positions.
\end{itemize}

\paragraph{Verifier.}
For each candidate $\mathcal{U}_t$, we define a look-ahead state $\tilde{x}_{t-1}$, 
from which the corresponding predictive distribution $p_\theta(\cdot \mid \tilde{x}_{t-1})$ 
is obtained. The verifier then evaluates this distribution and assigns an uncertainty
score reflecting the expected consistency with future predictions.
\[
\tilde{x}_{t-1}^i \sim p_\theta^i(\cdot \mid x_t), 
\;\; \forall i \in \mathcal{U}_t,
\qquad 
V(p_\theta(\cdot|\tilde{x}_{t-1}))\in \mathbb{R}.
\]
We propose to quantify uncertainty by leveraging the probability distribution generated by the model and employ this measure as the basis of the verifier. The verifier operates in intermediate sampling states, but evaluates them using a sequence-level approximation to the prediction of the model of $x_0$. The verifier operates in intermediate sampling states, quantifying the potential
certainty over the entire sequence of tokens. We propose several candidate functions:
\begin{itemize}[leftmargin=1.5em]
    \item \textbf{Average Negative Entropy:} Instantiate the verifier with the mean Shannon entropy of the predictive distribution induced by the model under the look-ahead state. To maintain consistency with the other verifiers and to interpret entropy as a certainty measure, we adopt the negative entropy, so that larger scores correspond to higher certainty. For each position $i\in [L]$, define 
\[
V(p_\theta(\cdot|\tilde{x}_{t-1}))=-\frac{1}{L}\sum_{i\in[L]}H(p_\theta^i(\cdot|\tilde{x}_{t-1})).
\]
Entropy aggregates information across the entire sequence distribution. This makes it 
sensitive not only to the dominance of the leading token but also to the dispersion of 
probability mass among lower-ranked alternatives.
    \item \textbf{Average Confidence:} Instantiate the verifier with the mean of the maximum probabilities of the predictive distributions under the look-ahead state. For each position $i\in [L]$, define
    \[
    V(p_\theta(\cdot|\tilde{x}_{t-1}))=\frac{1}{L}\sum_{i\in[L]}\max_{v\in\mathcal{V}}\ p_\theta^i(v|\tilde{x}_{t-1}).
    \]
Confidence reflects the degree to which the model concentrates probability mass on a single outcome, thereby serving as a straightforward indicator of positional certainty across the sequence.
\end{itemize}
The interaction between the path generator and the verifier can be naturally viewed as a two-stage sampling process: the generator proposes candidate unmasking paths, and the verifier scores these candidates based on predictive consistency. By preferentially selecting paths with lower uncertainty, this process aligns generation with more reliable trajectories, similar to how reward models guide sampling in reinforcement learning. From this perspective, the unmasking can be interpreted as a form of reward alignment, where the verifier acts as a surrogate reward model. We adopt two sampling schemes: Sequential Monte Carlo (SMC), which propagates weighted particles through the denoising process with incremental reweighting, and Nested Importance Sampling (NIS), which performs importance weighting at each step based on immediate reward estimates. These approaches follow established methodologies for reward-aligned sampling in previous work~\citep{li2024derivative,wu2023practical}. The concrete implementation details of both methods are provided in Appendix~\ref{appendix:method}.
\subsection{Algorithm Design and Analysis}
The proposed \algname{} combines path generation with uncertainty verification in a unified framework, thereby guiding the model away from erroneous unmasking paths. The design of the algorithm depends on the specific choice of verifier, path generator, and sampling scheme, while its computational complexity is determined by the number of candidate paths generated during each iteration. This section provides a formal design of the algorithm and analyzes its computational properties.

\paragraph{Algorithm.} Algorithm~\ref{alg:lookahead} illustrates the overall pseudocode procedure. Starting from a fully masked sequence $x_T$, the algorithm iteratively reduces the mask set $\mathcal{M}_t$ until a complete sequence $x_0$ is obtained. At each step $t$, the path generator $G$ produces $k$ candidate unmasking sets $\{\mathcal{U}_{t,i}\}_{i=1}^k$, which are sampled from a certainty-based pool $\mathcal{P}_t$ (lines 4-5). Each candidate defines a hypothetical next state $\tilde{x}_{t-1,i}$ by unmasking tokens at the selected positions according to their predictive distributions. The verifier $V$ then assigns uncertainty-based scores  $\{s_i\}_{i=1}^k$ to these states (line 6), and the actual next state $\tilde{x}_{t-1}$ follows NIS or SMC (line 7).

\begin{algorithm}[ht]
\caption{Algorithm of \algname{}}\label{alg:lookahead}
\centering
\begin{algorithmic}[1]
\State \textbf{Require:} Verifier $V$, path generator $G$, diffusion model $p_\theta$, budget for each timesteps $\{B_t\}_{t=T}^1$, lookahead number k.
\State \textbf{Init:} Masked sequence $x_T$
\For{$t \in \{T, \cdots, 1\}$}
    \State Model prediction with $p_\theta(\cdot|x_t)$ and make pool $\mathcal{P}_t$
    \State generate $k$ paths $\{\mathcal{U}_{t,i}\}_{i=1}^k$ with generator $G(p_\theta(\cdot|x_t),B_t,\mathcal{P}_t)$ and sample candidates $\{\tilde{x}_{t-1,i}\}_{i=1}^k$ using $\{\mathcal{U}_{t,i}\}_{i=1}^k$ \Comment{Path Generation}
    \State Evaluate future predictions $\{p_\theta(\cdot|\tilde{x}_{t-1,i})\}_{i=1}^k$ using verifier $V$ to obtain uncertainty scores $\{s_i\}_{i=1}^k$ \Comment{Verification}
    \State Select $\tilde{x}_{t-1}$ from $\{\tilde{x}_{t-1,i}\}_{i=1}^k$ using importance sampling (SMC/NIS) based on scores $\{s_i\}_{i=1}^k$ \Comment{Selection}
\EndFor
\end{algorithmic}
\end{algorithm}
\paragraph{Complexity Analysis.} 
\algname{}'s  computational overhead stems primarily from evaluating k candidate paths per denoising step. The verifier $V$ and pool construction $\mathcal{P}_t$  introduce negligible overhead, as they involve only tensor operations on precomputed logits without additional model forward passes. As shown in Figure~\ref{realscaling}, $k=2,3$ achieves near-optimal performance across benchmarks, yielding a 2-3× computational cost comparable to classifier-free guidance~\citep{ho2022classifier}. This efficient scaling makes LookUM practical for deployment while significantly improving generation quality (Section~\ref{analysis}).

\section{Experiments}
\label{mainexp}
\subsection{Experimental Setup}
\label{mainexpsetup}
\paragraph{Datasets.} We evaluate \algname{} on diverse reasoning benchmarks, including MATH500~\citep{lightman2023let} and GSM8K~\citep{cobbe2021training} for mathematics, HumanEval~\citep{chen2021evaluating} and MBPP~\citep{austin2021program} for coding, and Sudoku and Countdown for planning tasks.

\paragraph{Baselines.} We compare \algname{} with \textit{five} unmasking strategies: (1) Confidence-based unmasking~\citep{chang2022maskgit} that sequentially unmasks tokens in descending order of prediction confidence, (2) Margin-based unmasking~\citep{kim2025train} that unmasks tokens based on the margin between their top-1 and top-2 prediction probabilities, (3) Entropy-based unmasking~\citep{koh2024plm} that unmasks tokens in ascending order of entropy, and (4) PC-Sampler~\citep{huang2025pc} that calibrates the original confidence by incorporating position-aware weighting and frequency-based adjustment. In addition, although they are not Unmasking Strategies, we also experiment with (5) ReMDM~\citep{wang2025remasking} which dynamically remasks and regenerates low-confidence tokens during generation.

\paragraph{Implementation Details.} We use LLaDA-8B-Instruct~\citep{nie2025large} and LLaDA-1.5~\citep{zhu2025llada}  for main experiments. Following the decoding setup in \citep{zhao2025d1}, we set the sequence lengths to 128 and 256 and apply unmasking with two tokens per step. All experiments are conducted using two NVIDIA A100 GPUs.

\subsection{Main Results}
Table~\ref{tab:main} demonstrates that Lookahead Unmasking consistently outperforms baseline sampling strategies across both LLaDA-8B and the RL-tuned LLaDA 1.5, without requiring additional model fine-tuning. The method achieves substantial improvements: an absolute gain of 8 points on HumanEval at sequence length 128 and 4 points on GSM8K for LLaDA. Notably, on LLaDA 1.5, which has already undergone reinforcement learning optimization, LookUM still delivers significant gains. This demonstrates that our uncertainty-based approach captures complementary signals to reward-based training, as it can further enhance models that have already been optimized through RL.

The consistent improvements across both base and RL-tuned models provide strong empirical evidence that Lookahead Unmasking effectively avoids erroneous unmasking paths. Particularly striking is that LLaDA-8B with LookUM achieves competitive or superior performance to vanilla LLaDA 1.5 on several benchmarks, suggesting that inference-time optimization can partially substitute for expensive RL training.

While ReMDM attempts to mitigate local errors by remasking low-confidence tokens, its performance gains are inconsistent. Results on Countdown, HumanEval, and MBPP show minimal improvement or degradation, indicating that remasking may disrupt solution trajectories in tasks requiring global structural coherence. By preserving the natural generative path without altering established structure, Lookahead Unmasking delivers more reliable improvements across diverse benchmarks.

\setlength{\tabcolsep}{3pt} 
\begin{table}[t!]
\caption{\textbf{Performance comparison across models.} Results on six reasoning benchmarks with LLaDA-8B and LLaDA-1.5 (RL-tuned). Best values are in \colorbox{blue!10}{blue} and the second best values are in \colorbox{green!12}{green}.}
\vspace{10pt}
\centering
\renewcommand{\arraystretch}{1.2} 
\resizebox{\textwidth}{!}{%
\begin{tabular}{@{\hspace{3pt}}l@{\hspace{8pt}}l@{\hspace{12pt}}cc@{\hspace{10pt}}cc@{\hspace{10pt}}cc@{\hspace{10pt}}cc@{\hspace{10pt}}cc@{\hspace{10pt}}cc@{\hspace{3pt}}}
\toprule
\textbf{Model} & \textbf{Method} 
& \multicolumn{2}{c}{\textbf{MBPP}} 
& \multicolumn{2}{c}{\textbf{Humaneval}} 
& \multicolumn{2}{c}{\textbf{GSM8K}} 
& \multicolumn{2}{c}{\textbf{MATH500}} 
& \multicolumn{2}{c}{\textbf{Countdown}} 
& \multicolumn{2}{c}{\textbf{Sudoku}} \\
\cmidrule(lr){3-4} \cmidrule(lr){5-6} \cmidrule(lr){7-8} 
\cmidrule(lr){9-10} \cmidrule(lr){11-12} \cmidrule(lr){13-14}
& & 128 & 256 & 128 & 256 & 128 & 256 & 128 & 256 & 128 & 256 & 128 & 256 \\
\midrule
\multirow{6}{*}{\parbox{2cm}{LLaDA}} 
& Confidence       & \colorbox{green!12}{28.6} & \colorbox{green!12}{28.4} & 19.5 & 32.0 & 68.3 & 76.7 & 26.0 & 32.4 & 20.3 & \colorbox{green!12}{21.9} & 1.4 & 27.4 \\
& Margin          & \colorbox{green!12}{28.6} &\colorbox{green!12}{28.4} & \colorbox{green!12}{25.6} & 31.0 & 67.1 & 76.1 & \colorbox{green!12}{28.4} & \colorbox{green!12}{34.4} & 19.1 & 20.7 & 21.8 & \colorbox{green!12}{27.8} \\
& Entropy         & 27.2 & 24.4 & 19.5 & 15.9 & 66.7 & 75.4 & 26.0 & 33.0 & 21.9 & 20.3 & 0.0 & 12.0 \\
& PC-Sampler      & 24.0 & 25.2 & 13.4 & 
\colorbox{green!12}{30.5} & 67.3 & 73.7 & 25.2 & 32.4 & \colorbox{blue!10}{26.5} & 20.3 & \colorbox{green!12}{23.2} & 24.0 \\
& ReMDM           & 28.6 & \colorbox{green!12}{28.4} & 17.1 & 29.3 & \colorbox{green!12}{69.1} &\colorbox{green!12}{77.9} & 27.4 & 33.0 & 25.3 & 17.2 & 0.4 & 22.8 \\
& \textbf{LookUM} & \colorbox{blue!10}{30.5} & \colorbox{blue!10}{36.2} & \colorbox{blue!10}{27.4} & \colorbox{blue!10}{35.9} & \colorbox{blue!10}{72.7} & \colorbox{blue!10}{79.3} & \colorbox{blue!10}{28.8} & \colorbox{blue!10}{34.6} & \colorbox{green!12}{25.4} & \colorbox{blue!10}{23.1} & \colorbox{blue!10}{25.0} & \colorbox{blue!10}{28.0} \\
\midrule[0.8pt]
\multirow{6}{*}{\parbox{2cm}{LLaDA-1.5}} 
& Confidence       & 40.3 & 38.6 & 26.8 & 23.7 & 69.5 & 79.4 & 28.6 & 32.6 & 20.3 & \colorbox{blue!10}{23.4}& 1.4 & 27.4 \\
& Margin          & 39.5 & 38.1 & \colorbox{blue!10}{31.1} & 27.4 & \colorbox{green!12}{71.3} & 78.3 & 27.2 & \colorbox{green!12}{35.0} & 24.6 & 14.0 & 21.8 &\colorbox{green!12}{27.8} \\
& Entropy         & 40.7 & 36.5 & 20.7 & 23.2 & 69.7 & 77.0 & 28.2 & 32.2 & 23.0 & 12.9 & 0.0 & 12.0 \\
& PC-Sampler      & \colorbox{green!12}{42.8} & \colorbox{green!12}{39.6} & 23.8 & 23.8 & 70.1 & 77.3 & 26.6 & 32.2 & \colorbox{green!12}{25.4} & 19.1 & \colorbox{green!12}{25.6} & 27.2 \\
& ReMDM           & 41.9 & 39.3 & 28.1 & \colorbox{green!12}{29.8} & 70.4 & \colorbox{green!12}{80.1} & 27.4 & 34.0 & 23.4 & \colorbox{green!12}{19.9} & 0.4 & 22.8 \\
& \textbf{LookUM} & \colorbox{blue!10}{45.0} & \colorbox{blue!10}{43.6} & \colorbox{green!12}{30.7} & \colorbox{blue!10}{33.5} & \colorbox{blue!10}{74.5} & \colorbox{blue!10}{82.3} & \colorbox{blue!10}{29.2} & \colorbox{blue!10}{35.8} & \colorbox{blue!10}{27.3} & 17.9 & \colorbox{blue!10}{26.8} & \colorbox{blue!10}{28.0} \\
\bottomrule
\end{tabular}
}
\label{tab:main}
\end{table}
\begin{figure*}[ht!]
    \centering
    \includegraphics[width=\linewidth]{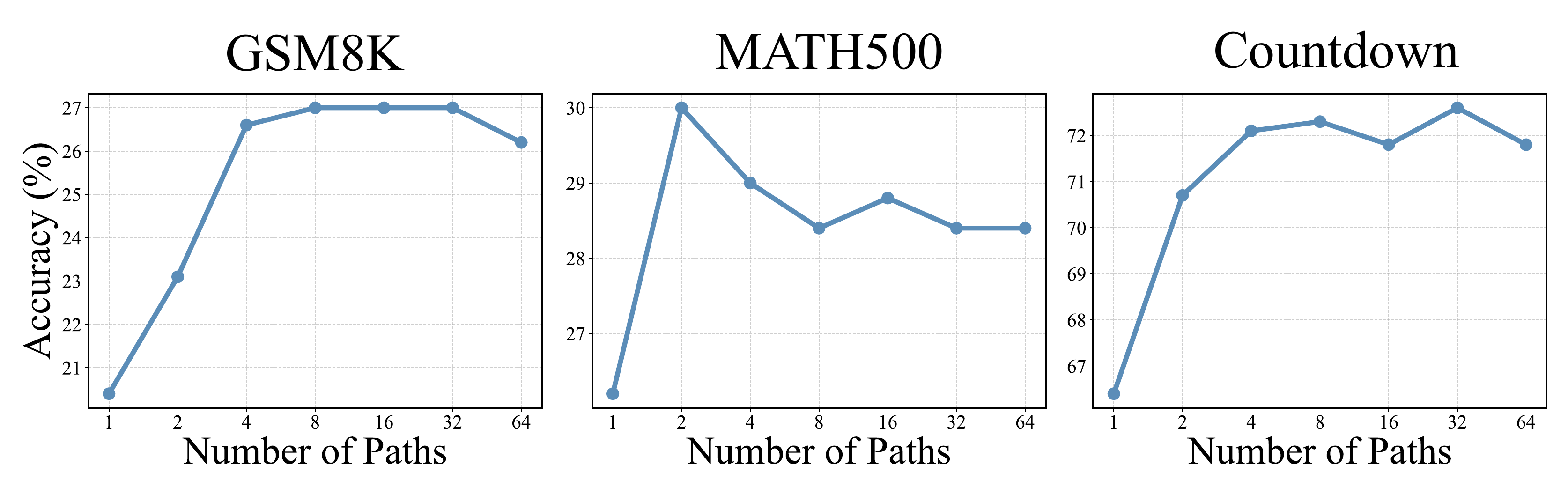}
    \vspace{-20pt}
    \caption{\textbf{Scaling results.} Performance scaling with lookahead paths. Accuracy versus number of particle paths on three benchmarks (GSM8K, MATH500, Countdown). Sharp improvements occur up to 4 particles, after which performance saturates, demonstrating efficient scaling with limited computational budget.}
    \label{realscaling}
\end{figure*}
\subsection{Analysis}
\label{analysis}
\paragraph{Experiments with External Reward Model.}
A key advantage of \algname{} is its independence from external reward models, eliminating the need for costly task-specific training or model deployment. To validate this design choice, we investigate whether incorporating mathematical process reward models could enhance performance. We employ Qwen2.5-Math-PRM-7B~\citep{zhang2025lessons} as an alternative verifier and evaluate it using both NIS and SMC sampling methods with varying numbers of particles (1, 4, 16) on MATH500 and GSM8K.

Table~\ref{tab:rm} shows reward-based verification underperforms our uncertainty-based approach. Reward models expect coherent partial solutions but encounter noisy~\citep{gong2025diffucoder}, globally-distributed predictions in intermediate states of DLMs. This incompatibility validates our use of intrinsic uncertainty signals, confirming that model-free design of \algname{} is more effective and practical for diffusion language models.

\begin{wraptable}{r}{0.42 \linewidth}
\centering
\vspace{-0.7em}
\caption{\textbf{Performance with external reward models.}}

\vspace{10pt}
\begin{tabular}{lccc ccc}
\toprule
\textbf{PRM}
& \multicolumn{3}{c}{\textbf{MATH500}}
& \multicolumn{3}{c}{\textbf{GSM8K}}
\\
\cmidrule(lr){2-4} \cmidrule(lr){5-7}
& \textbf{1} & \textbf{4} & \textbf{16}
& \textbf{1} & \textbf{4} & \textbf{16}

 \\
\midrule
NIS     & 26.0& 26.6 & 26.0& 69.0& 69.1 & 69.5 \\
SMC     & 26.0 & 22.6 & 24.0 & 69.0 & 68.0 & 68.5  \\
\bottomrule

\end{tabular}
\vspace{-0.5em}
\label{tab:rm}
\end{wraptable}

\paragraph{Inference Time Scaling.}
The SMC-based process \algname{} is extensible with respect to the number of particles used. We regard the number of particles as the scaling axis and investigate the inference-time scaling effect. We use the \emph{number of paths} generated by the path generator as our measure of computational cost, as each path requires a separate model evaluation while the  computational overhead of verifier remains negligible. The experiments are conducted with LLaDA on GSM8K, MATH500, and Countdown, with the generation length fixed at 128 and the number of paths set to $k\in \{1,2,\ldots,64\}$.

As illustrated in Figure~\ref{realscaling}, we observe distinct scaling patterns across benchmarks. GSM8K shows substantial improvements up to 4 paths, after which performance plateaus. MATH500 achieves optimal performance with just 2 paths, with additional paths leading to slight performance degradation. These results reveal that \algname{} reaches optimal performance with merely 2-4 paths, requiring comparable computational overhead to classifier-free guidance. Rapid saturation indicates that our uncertainty-based verifier efficiently identifies promising unmasking trajectories without exhaustive search, making the method both practical and robust for deployment under computational constraints.

\subsection{Component Discovery}
\begin{table}[h]
\centering
\caption{\textbf{Ablation study} of \algname{} components on GSM8K, MATH500 and Countdown.}
\vspace{10pt}
\label{tab:ablation}
\begin{tabular}{l@{\hspace{1cm}}lc@{\hspace{0.7cm}}c@{\hspace{0.7cm}}c}
\toprule
Component & Variant & GSM8K & MATH500 &Countdown \\
\midrule
\multirow{4}{*}{Path Generator} & N-best (Confidence) &72.1  &\textbf{29.0}  &\textbf{32.0}  \\
& N-best (Margin) & 72.3 &28.6  &31.3  \\
& N-best (Negative Entropy) &72.5  &28.8  &28.1\\
 & Certainty Filtering &\textbf{72.6} &26.4  &19.5  \\
 
\midrule
\multirow{2}{*}{Verifier} & Avg. Negative Entropy &\textbf{72.3} &\textbf{28.6}  &\textbf{31.3}   \\
 & Avg. Confidence & 71.0 &  27.8& 30.5  \\
\midrule
\multirow{2}{*}{Sampling} & SMC &70.7  &\textbf{29.0}  &23.1   \\
 & NIS &\textbf{72.3} &28.6  &\textbf{31.3}  \\
\bottomrule
\end{tabular}
\end{table}
We conduct ablation studies to analyze the contribution of each component in \algname{}: the path generator, verifier, and sampling method. Table \ref{tab:ablation} presents results on GSM8K, MATH500, and Countdown with sequence length 128 and 64 denoising steps, varying one component at a time while keeping others fixed. Full experimental details are provided in Appendix~\ref{expdetail}.
\paragraph{Path Generator.} We evaluate four pooling strategies: N-best with confidence, margin, and entropy criteria, and certainty filtering. Results show that N-best pooling variants achieve comparable performance on GSM8K, while certainty filtering yields the highest accuracy on GSM8K but degrades significantly on MATH500 and Countdown. This suggests that adaptive threshold-based selection is less robust across different task complexities than rank-based selection.

\paragraph{Verifier.} The comparison between average negative entropy and average confidence verifiers reveals the importance of capturing distributional uncertainty beyond just the top token. Average negative entropy consistently outperforms confidence across all benchmarks. This performance difference stems from their fundamental measurement properties: while confidence only examines the maximum probability at each position, entropy captures the complete shape of the probability distribution, including the dispersion among lower-ranked tokens.

\paragraph{Sampling Method.}
The choice between NIS and SMC reveals task-dependent trade-offs in how uncertainty signals should be propagated through the denoising process. NIS, which performs independent importance weighting at each step, excels on GSM8K and shows particularly striking advantages on Countdown. This suggests that for tasks with strong local structure—where each step's correctness can be evaluated relatively independently—the immediate feedback of NIS effectively steers generation away from errors.

These ablation results reveal that \algname{} effectiveness stems not from any single component but from the synergistic interaction between thoughtfully designed elements. The path generator ensures diverse yet high-quality candidates, the verifier provides rich uncertainty signals that capture both local and global coherence, and the sampling method determines how these signals guide the denoising trajectory.


\section{Related Work}
\paragraph{Unmasking Strategies for Masked Diffusion Models.}
MDMs reconstruct sequences through iterative predictions during the Unmasking process. The order-agnostic training of MDMs permits diverse unmasking paths during sampling, and a variety of strategies have been proposed to exploit this flexibility for performance improvement. A baseline approach selects tokens for unmasking at random, consistent with the training procedure, but this strategy often leads to performance degradation in specific tasks. The High-Confidence Unmasking strategy~\citep{chang2022maskgit} unmask tokens with the highest confidence in the intermediate prediction of the model, and it has been employed in LLaDA. The Top Probability Margin strategy~\citep{kim2025train} selects tokens for unmasking based on the difference between the two highest probability values at each position, and it has demonstrated substantial performance gains in logic puzzles. Dream further introduces a strategy that unmask tokens at positions with low entropy. PC-Sampler~\citep{huang2025pc} calibrates confidence scores using position-aware weighting and token frequency distribution to address the model biases. However, these methods all rely solely on token-level local scores, failing to capture dependencies between tokens to be unmasked. ReMDM~\citep{wang2025remasking} attempts to correct errors by dynamically remasking previously unmasked tokens during intermediate generation steps.

\paragraph{Reward Guided Generation.}
Reward Guided Generation has been explored in several studies, including works on protein design that employ approaches such as SMC~\citep{doucet2001introduction}, SVDD~\citep{li2024derivative} and FK Steering~\citep{singhal2025general}. Concretely, these approaches guide the denoising process toward high-reward regions through distinct mechanisms: SMC-based methods maintain ensembles of particles throughout generation, iteratively re-weighting them to concentrate on promising trajectories, whereas SVDD performs stepwise path selection by generating multiple candidates at each stage and choosing the highest-scoring continuation. However, these approaches universally require task-specific reward models that must be trained separately, not only limiting scalability but also rendering them impractical for complex reasoning tasks such as mathematical problem-solving and code generation. In contrast, our method evaluates the model predictions via an intrinsic uncertainty-oriented function, thus obviating the reliance on an external reward model.

\section{Conclusion}
We proposed Lookahead Unmasking, an unsupervised inference-time framework that reformulates unmasking as a path selection problem guided by uncertainty. Lookahead Unmasking effectively reduces local errors, scales with additional compute, and consistently improves reasoning performance across mathematics, coding, and planning benchmarks, without relying on external reward models. Our results demonstrate that uncertainty-aware unmasking offers a simple and general approach for advancing diffusion language models.

Future work might fruitfully investigate verifiers that examine intermediate model representations beyond output probabilities, particularly attention patterns across layers and timesteps. Such exploration could potentially reveal which token relationships the model prioritizes during unmasking, possibly offering signals orthogonal to uncertainty for path selection. We speculate that combining multiple intrinsic signals (uncertainty, attention weights, gradient magnitudes) could lead to more robust verifiers that may better capture the internal reasoning process of the model. Exploring these internal dynamics might open new avenues for sophisticated path selection strategies that could extend beyond surface level output analysis.
\section*{Reproducibility Statement}
We provide hyperparameter details and setup of all experiments in Section~\ref{mainexpsetup} and  Appendix~\ref{expdetail}.
\bibliography{main.bib}
\bibliographystyle{iclr2026_conference}
\newpage
\appendix
\section*{\Large Appendix}

\section{Limitations}
 While our work demonstrates that uncertainty provides a strong signal for path selection, other intrinsic model signals might offer complementary benefits. Our \algname{} suggests that uncertainty-based verifiers may be particularly well suited for evaluating the noisy intermediate states characteristic of diffusion models, though this approach relies solely on model output distributions. In continuous diffusion domains, attention manipulation has shown promise for guiding generation~\citep{ahn2024self}, yet the potential for leveraging such internal representations in discrete diffusion remains largely unexplored. 

\section{Sampling Algorithm for Lookahead Unmasking}
\label{appendix:method}
\subsection{Importance Sampling}
Importance Samplings are widely employed to approximate expectations of complex 
functions with respect to a probability distribution. Specifically, given a target 
distribution $p^\ast(x)$, the goal is to compute
\begin{equation}
    \mu = \mathbb{E}_{p^\ast}[f(x)] = \int f(x)\, p(x)\, dx .
\end{equation}

In many cases direct sampling from $p^\ast(x)$ is difficult, or 
rare events of interest occur with extremely low probability. Which
alternative proposal distribution $p(x)$, from which sampling is easier, 
and reweight the samples accordingly:
\begin{equation}
    \mu = \int f(x)\, p(x)\, dx 
        = \int f(x)\, \frac{p^\ast(x)}{p(x)} \, p(x)\, dx 
        = \mathbb{E}_{p}[ f(x)\, w(x) ],
\end{equation}
where $w(x) = \frac{p(x)}{q(x)}$ is referred to as the 
\emph{importance weight}.
By choosing $f$ as a Dirac-delta function, the procedure can be viewed as an instance of important resampling. In this setting, normalized weights enable sampling that approximates draws from $p^\ast(x)$. 
The pseudocode of Important Sampling is presented as follows:
\begin{algorithm}[H]
\caption{Importance Sampling}
\begin{algorithmic}[1]
\Require Target distribution $p^\ast(x)$, proposal distribution $p(x)$, function $f(x)$, number of samples $N$
\For{$i = 1$ to $N$}
    \State Sample $x_i \sim p(x)$
    \State Compute weight $w_i \gets \frac{p^\ast(x_i)}{p(x_i)}$
\EndFor
\State Normalize weights: $\tilde{w}_i \gets \dfrac{w_i}{\sum_{j=1}^N w_j}$
\State Sample $x \sim \sum_{i=1}^N \tilde{w}_i \, \delta_{x_i}$
\end{algorithmic}
\end{algorithm}

\subsection{Sequential Monte Carlo}
Sequential Monte Carlo (SMC), often referred to as particle filtering,
is a family of algorithms for approximating sequences of probability
distributions like diffusion models.
The idea is to propagate a set of weighted samples through
time using importance sampling and resampling steps.
The algorithm SMC is as follows:
\begin{algorithm}[H]
\caption{Sequential Monte Carlo}
\begin{algorithmic}[1]
\State \textbf{Require:} number of particles $N$, initial distribution $p(x_0)$, proposal kernels $p(\cdot \mid x_{t-1})$, target kernels $p^\ast(\cdot \mid x_{t-1})$
\State \textbf{Initialization:}
\For{$i = 1,\dots,N$}
    \State Draw $N$ samples $x_0^{(i)} \sim p(x_0)$
    \State Set weight $w_0^{(i)} =1$
\EndFor

\For{$t = 1,\dots,T$}
    \State \textbf{Propagation:} Draw $N$ samples $x_t^{(i)} \sim q_t(\cdot \mid x_{t-1}^{(i)})$
    \State \textbf{Weighting:}
    \[
    w_t^{(i)} 
      =\frac{p^\ast_t(x_{t}^{(i)})}
           {p_{t-1}^\ast(x_{t-1}^{(i)})\, p_t(x_t^{(i)} \mid x_{t-1}^{(i)})}
    \]
    \State Normalize $\tilde w_t^{(i)} = w_t^{(i)} / \sum_{j=1}^N w_t^{(j)}$
    \State \textbf{Resample:} Replace $\{x_t^{(i)}\}_{i=1}^N$ with indices drawn 
        according to $\tilde w_t^{(1:N)}$.
\EndFor
\State \textbf{Output:} weighted particle system $\{x_{0:T}^{(i)}, \tilde w_T^{(i)}\}_{i=1}^N$
\end{algorithmic}
\end{algorithm}

\subsection{Reward Guided Sampling}
The objective of reward-aligned sampling is to maintain the naturalness of generated samples while optimizing for the specified reward, or more broadly, to generate samples from a target distribution that encodes the underlying preferences. Concretely, the objective of reward-guided sampling with a reward function $r(\cdot)$ is to draw samples from the target distribution:
\[
p^{\ast}(\cdot) = \arg\max_{p} \, 
\mathbb{E}_{x_0 \sim p(\cdot)} \big[ r(x_0) \big] 
- \alpha \, \KL\left( p(x_0) \Vert p(x_0) \right)\propto \exp\!\big(r(\cdot)/\alpha) p_\theta(\cdot)
\]
where $\alpha$ controls the strength of the KL divergence regularization term. The reward is a measure of how well the model output matches the desired criteria. It can be broadly defined to include classifier evaluations, correctness, and other factors. We consider the certainty score obtained from a verifier as the reward. We consider the certainty score obtained from a verifier as the reward. The verifier evaluates the prediction of the model for the entire sequence to produce this score.

A diffusion model generates samples sequentially and each transition kernel must be optimized to enable sampling from the target distribution. When the intermediate reward is defined as the expectation of future rewards, the optimal transition kernel can be derived as follows:
\[
p^{\ast}(\cdot | x_t) \propto p_\theta(\cdot | x_t) \exp(r(\cdot)/\alpha).
\]
By performing denoising with the optimal transition kernel, each intermediate distribution can be expressed as follows:
\[
p^\ast(x_t)=\frac{p(x_t)\exp{r(x_t)/\alpha}}{\sum p(x_t)\exp{r(x_t)/\alpha}}.
\]

Existing approaches to approximate the optimal transition kernel using Sequential Monte Carlo~\citep{singhal2025general} or Nested Importance Sampling~\citep{li2024derivative}. Nested Importance Sampling(NIS) is a method that estimates expectations by applying importance sampling at every stage of a sequential process. The weight $w_t^{(i)}$ for each sampling step is given by:
\[
\text{NIS: }w_t^{(i)}= \exp{(r(x_t^{(i)})/\alpha)} \quad \text{SMC: } w_t^{(i)}=\exp((r(x_{t-1})-r(x_t))/\alpha),
\]
where the generative process is defined in reverse time, in a manner that is compatible with the denoising process.
\section{Experimental Details}
\label{expdetail}
\subsection{Experimental Details in Section~\ref{localerror}}
\paragraph{Table~\ref{tab:uncertainty}.}
We use arithmetic datasets (subtraction, addition, multiplication) with 2,000 two-number calculations each. Generation length is set to 8 tokens, with arithmetic questions serving as prompts. To analyze local error impact, we use GPT-4o to create erroneous versions by introducing computational mistakes, then compare mean negative entropy and mean confidence between correct and error-containing trajectories. This controlled setup quantifies how local errors affect model uncertainty during generation.

\paragraph{Figure~\ref{fig:local_error}.}
The generation was performed with the same setup as the main experiment, using LLaDA-8B-Instruct with a sequence length of 128 and 64 steps, and a temperature of 0.1. For Lookahead Sampling, we set $|\mathcal{P}_t|=5$, selected two paths, and verify samples using Average Negative Entropy. We measure the error rate by counting local errors at the sentence level using GPT-4o. The system prompt is configured to strictly evaluate only local errors at the sentence level. The experimental results are presented in Figure~\ref{fig:local_error}.

\subsection{Experimental Details in Section~\ref{mainexp}}
\paragraph{Evaluation Hyperparameters.}
In the experiments, we evaluated two types of DLMs: LLaDA, Dream, and LLaDA-1.5. For all two experiments, Lookahead Sampling was performed with 
$|\mathcal{P}_t|=5$, two selected paths, and the Average Negative Entropy setting. The sequence lengths were set to 128 and 256, and unmasking with two tokens per step was applied in all cases. The baselines, ReMDM and PC-Sampler, were implemented following the original implementations provided in their respective papers.

\paragraph{With Externel Reward Model.}  We consider  Qwen2.5-Math-PRM-7B as the reward model. The experiments are conducted with LLaDA, and at each step we apply the Resample stage of SMC and NIS. At every step, we evaluate the prediction of $x_0$ and adopt $\alpha=0.1$, Confidence Unmasking which is commonly used in prior work.

\paragraph{Ablation} 
We conduct an ablation study by varying the path generator, verifier, and sampling method while keeping the evaluation setting as the default. All experiments are performed with a sequence length of 128, temperature 0.2, $|\mathcal{P}_t|=5$ and the results are compared on MATH500,GSM8K and Countdown benchmarks.
\end{document}